\documentclass[]{spie}  

\usepackage[utf8]{inputenc}
\usepackage{caption}
\usepackage{amsfonts}
\usepackage{setspace}
\usepackage{algorithm}
\usepackage{algpseudocode}
\DeclareCaptionLabelSeparator{twospace}{~~~~}
\usepackage{indentfirst}
\usepackage{hyperref}
\hypersetup{colorlinks,breaklinks,
             linkcolor=blue,urlcolor=blue,
             anchorcolor=blue,citecolor=blue}

\usepackage{graphicx,amsmath}
\usepackage{subcaption}
\usepackage{float}

\usepackage{todonotes}
\usepackage{multirow}

\usepackage{fourier} 
\usepackage{array}
\usepackage{makecell}
\usepackage{eucal}
\usepackage{enumerate}

\usepackage{threeparttable}
\usepackage{array} 
\usepackage{caption}
\usepackage{microtype}
\usepackage{chngcntr}

\usepackage{authblk}

\DeclareMathOperator*{\argmax}{arg\,max}
\DeclareMathOperator*{\argmin}{arg\,min}

\title{Novel Batch Active Learning Approach and Its Application to Synthetic Aperture Radar Datasets}
\author[1, a]{James Chapman}
\author[1, a]{ Bohan Chen}
\author[2, a]{ Zheng Tan}
\author[3, b]{\\Jeff Calder}
\author[3, a]{ Kevin Miller}
\author[3, a]{ Andrea L. Bertozzi}
\affil[a]{University of California, Los Angeles, Department of Mathematics, 520 Portola Plaza, Los Angeles, CA 90095, USA}
\affil[b]{University of Minnesota, School of Mathematics, 538 Vincent Hall, 206 Church Street SE, Minneapolis, MN 55455, USA}

\begin{document}

\maketitle

\section{Abstract}
Active learning improves the performance of machine learning methods by judiciously selecting a limited number of unlabeled data points to query for labels, with the aim of maximally improving the underlying classifier’s performance. Recent gains have been made using sequential active learning for synthetic aperture radar (SAR) data \cite{miller2022graph}. In each iteration, sequential active learning selects a query set of size one while batch active learning selects a query set of multiple datapoints. While batch active learning methods exhibit greater efficiency, the challenge lies in maintaining model accuracy relative to sequential active learning methods. We developed a novel, two-part approach for batch active learning: Dijkstra's Annulus Core-Set (DAC) for core-set generation and LocalMax for batch sampling. The batch active learning process that combines DAC and LocalMax achieves nearly identical accuracy as sequential active learning but is more efficient, proportional to the batch size. As an application, a pipeline is built based on transfer learning feature embedding, graph learning, DAC, and LocalMax to classify the FUSAR-Ship and OpenSARShip datasets. Our pipeline outperforms the state-of-the-art CNN-based methods.\footnote{Source Code: \url{https://github.com/chapman20j/SAR_BAL}\\
Please direct all correspondence to James Chapman: chapman20j@math.ucla.edu}

\section{Introduction}
\label{sec:intro}
Synthetic Aperture Radar (SAR) is a valuable tool in Automatic Target Recognition (ATR). SAR imaging uses a moving device, which repeatedly transmits and receives radio signals, to simulate a large radar dish and achieve high resolution images. Multiple standard SAR data sets have been established as benchmark data sets within the SAR community. For example, MSTAR contains SAR images of land based vehicles\cite{MSTAR}, whereas OpenSARShip and FUSAR-Ship contain SAR images of different types of ships at sea\cite{huang2017opensarship, hou2020fusar}. The main problem here is to identify objects from SAR images. Although the images are of higher resolution, they are difficult to classify via human inspection, often requiring lots of time from domain experts. This motivates the use of machine learning to improve speed and accuracy for object classification in SAR images. 

Due to its recent success in SAR classification and it's high data efficiency, we propose to use semi-supervised learning (SSL) for SAR classification\cite{miller2022graph}. Algorithms in this domain exploit the geometric structure of the entire dataset in addition to the small amount of label information. In particular, we use graph-based Laplace learning as the underlying classifier. In order to exploit the geometry of the SAR data, the methods must first obtain useful features from the images for making predictions. Convolutional neural network (CNN) architectures have been highly successful on image classification tasks as they learn rich, structured representations of image data. They are also equipped to deal with the large amounts of noise contained in SAR images. Past works have successfully used Convolutional Variational Auto-Encoders (CNNVAE) to embed SAR data into a lower dimensional embedding space \cite{miller2022graph}. Transfer learning presents another promising method for obtaining useful image features\cite{bansal2021transfer, arnold2018blending, inkawhich2021bridging}. This involves training a neural network on similar data for which labels are much more readily available. Later convolutional layers in the neural network contain image features useful for the prior task and may be used for SAR classification. These features can be used immediately, or may be fine tuned to get more task specific features after replacing the fully connected layers by linear layers and training on the current dataset. We employ CNNVAE, transfer learning without fine tuning and transfer learning with fine tuning in this paper.

While SSL methods perform relatively well with few labeled data, their performance drastically improves when combined with active learning\cite{zhu2005semi}. Active learning supports the machine learning process by judiciously selecting a small number of unlabeled datapoints to query for labels, with the aim of maximally improving the underlying classifier’s performance \cite{settles_active_2012}. Active learning has been shown to significantly improve classifier performance at very low label rates and minimize the cost of labeling data by domain experts \cite{miller2022graph, dasgupta_two_2011, miller2021model, settles_active_2012}. Central to active learning is the development of an acquisition function. This function quantifies the benefit of acquiring the label of each datapoint in the candidate set $\mathcal{C}$ (a subset of the unlabeled datapoints). Based on the acquisition function, the active learning process selects a query set $\mathcal{Q}\subset\mathcal{C}$ to label. Sequential active learning selects the maximizer of the acquisition function 
$$k^\ast = \argmax_{k\in\mathcal{C}}\ \mathcal{A}(k),$$
where "sequential" refers to the case in which the query set has size one (i.e. $|\mathcal{Q}|=1$). Batch active learning was developed to select a non-singleton query set in each step of the active learning process (i.e. $|\mathcal{Q}| > 1$). Batch active learning is more challenging than the sequential case as datapoints which maximize sequential acquisition functions usually contain similar information. Merely mimicking the sequential scheme to choose the top $|\mathcal{Q}|$ maximizers of the acquisition function $\mathcal{A}(k)$ is usually not optimal, since these maximizers are often close in the embedding space. Batch active learning methods remedy this issue by enforcing diversity in the query set, either by adding restrictions to sequential active learning or by designing objectives specific to batch active learning \cite{ji2012variance, ma2013sigma, cai2013maximizing, gal2017deep, kushnir2020diffusion, zhdanov2019diverse}. We opt for the former strategy and build on sequential acquisition functions. Later experiments compare our method, LocalMax, to other active learning methods. 

Recent works in ATR fall under two main approaches: supervised learning and SSL. Deep learning has enabled many recent advances as it can obtain useful image features. Zhang, et al. developed a supervised deep learning model called Hog-ShipCLSNet to classify the OpenSAR Ship and FUSAR-Ship data sets\cite{9445223}. Other notable supervised learning methods use transfer learning from simulated SAR datasets \cite{inkawhich2021bridging, arnold2018blending}. Simulating SAR data reduces the labeling burden on experts, but presents serious challenges due to distribution shift in the datasets \cite{lewis2019sar}. A notable SSL contribution was made in Miller, et al. where the authors applied a CNNVAE and graph-based sequential active learning to classify images in MSTAR\cite{miller2022graph}. This result achieved state of the art performance on the MSTAR dataset. 

\begin{figure}
    \centering
    \includegraphics[width=0.3\textwidth]{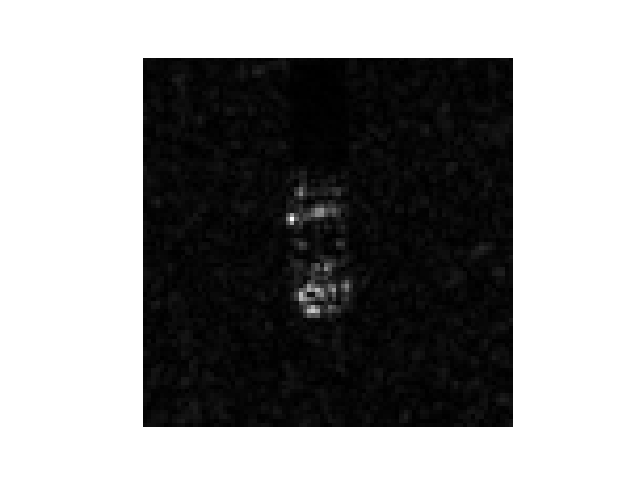}
    \includegraphics[width=0.3\textwidth]{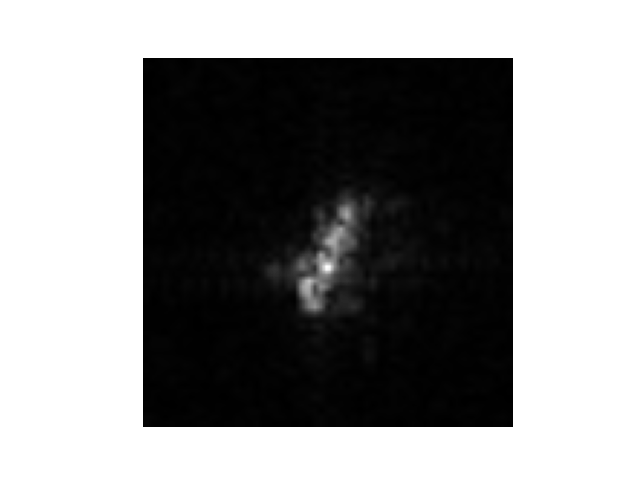}
    \includegraphics[width=0.3\textwidth]{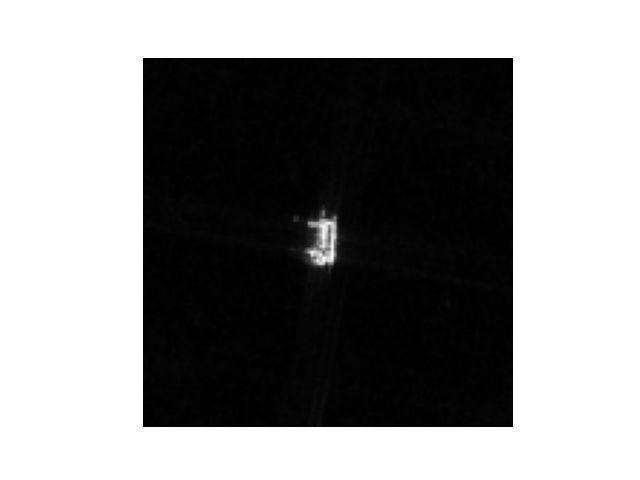}
    \caption{Three samples of SAR images. From left to right, the images show vehicles from MSTAR\cite{MSTAR}, OpenSARShip\cite{huang2017opensarship} and FUSAR-Ship\cite{hou2020fusar}.}
    \label{fig:sar_images}
\end{figure}

In this paper, we improve the feature embedding process on OpenSARShip and FUSAR-Ship using transfer learning from neural networks pre-trained on ImageNet \cite{ILSVRC15}. We also present a novel core-set construction method, DAC, and batch active learning procedure, LocalMax, to decrease the human labeling time by an order of magnitude. This method is compared with other active learning methods, including acquisition-weighted sampling, sequential active learning, random sampling and naive batch sampling. The combination of transfer learning and batch active learning beats the current state of the art methods on OpenSARShip and FUSAR-Ship, while simultaneously using less training data than state of the art methods \cite{9445223}. Results are also compared on the MSTAR dataset as a standard benchmark.

\newcommand{\y}{\mathbf{y}}
\newcommand{\bfu}{\mathbf{u}}
\newcommand{\ca}{C_\hat{A}}

\section{Math Background}
\label{sec:mathbg}
This section reviews some mathematical tools used in our pipeline, including CNNVAE, transfer learning,  $K$-nearest neighbor similarity graph construction, graph Laplace learning and active learning. The pipeline starts by embedding the data using either a CNNVAE or transfer learning.  The next step in the pipeline constructs a similarity graph based on the data embedding. Following this, the DAC algorithm computes a  core-set. Finally, the active learning loop cycles through: fitting the model with Laplace learning, selecting new points for labeling with LocalMax and querying the human for labels. The full pipeline is shown in Figure~\ref{fig:al_pipeline}.

\subsection{Data Embeddings}
\label{sec:embedding}
\begin{figure}
    \centering
    \includegraphics[width=0.9\textwidth]{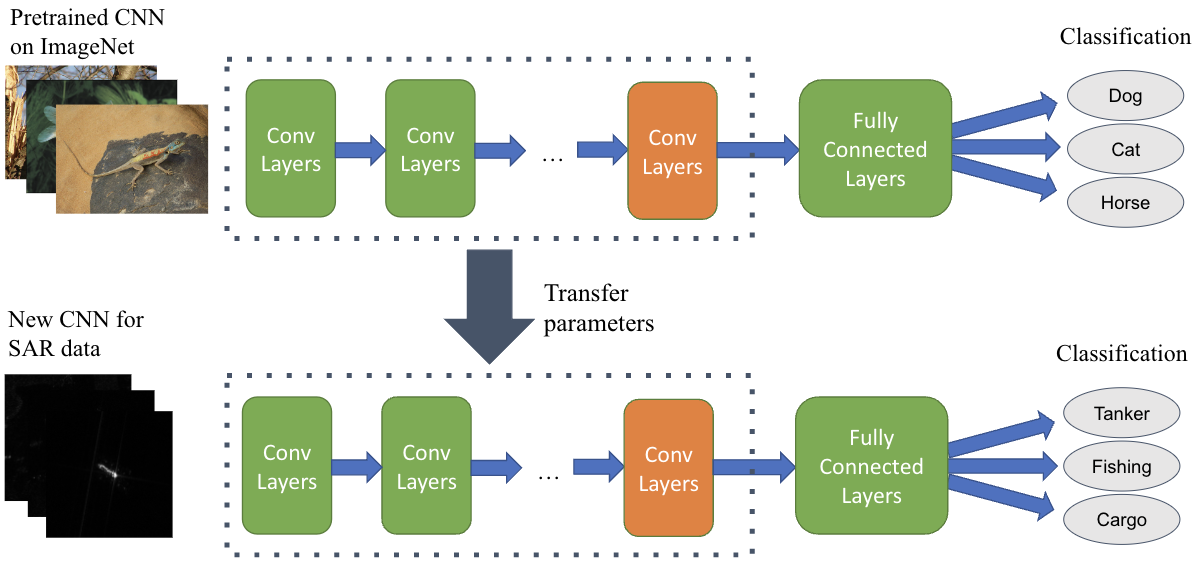}
    \caption{Flowchart for fine-tuned transfer learning. The parameters in the convolutional layers (contained in dotted boxes) of the pretrained CNN are transferred to the new CNN. In fine-tuned transfer learning, all the transferred parameters are trained for a few iterations on the new dataset. The training occurs by first adding new fully connected layers at the end of the neural network and performing supervised learning with the new dataset. When training is complete, only the layers in the dotted boxes are kept for the embedding process. The orange layer denotes the feature layer, and the outputs of this layer provide the feature vectors used later in the pipeline. }
    \label{fig:embedding_pipeline}
\end{figure}

As mentioned in Section~\ref{sec:intro}, our semi-supervised learning methods depend on a good representation of data, where distances between data measure similarity between those data. CNN architectures provide a good way to process image data. We utilize CNNVAEs from previous work\cite{miller2022graph} and use transfer learning from pre-trained PyTorch CNN neural networks to process the SAR datasets. A convolutional layer near the end of the neural network is designated as the \textit{feature layer} since it captures complex features of the dataset. The outputs of the feature layer are called \textit{feature vectors}; these are used in our subsequent graph construction where graph-based learning will take place. The transfer learning case is shown in Figure~\ref{fig:embedding_pipeline}, where the orange layers denote feature layers. 

Transfer learning works by reusing a neural network trained on a similar dataset and task for a new task. One strategy is to immediately use the neural network parameters from a pretrained neural network. Another strategy is to fine tune the parameters of the original neural network with respect to the new dataset. We denote these methods by \textit{zero-shot} transfer learning and \textit{fine-tuned} transfer learning, respectively. Fine-tuned transfer learning is shown with greater detail in Figure~\ref{fig:embedding_pipeline}. Alternatively, CNNVAEs work by first encoding and then decoding the dataset. The neural network architecture is designed so that the encoding is into a lower dimensional space. This forces the neural network to learn useful image features so that it can reconstruct the original image. In this method, the feature layer is chosen as the last CNN layer in the encoder portion of the CNNVAE. 

\subsection{Graph Construction}\label{sec:graph_construction}
As mentioned in Section~\ref{sec:intro}, our methods operate on a graph constructed from the data.
Consider the set of $d$-dimensional feature vectors $X = \{x_1,x_2,\ldots,x_N\}\subset \mathbb{R}^d$. We build a graph $G=(X,W)$, where $X$ is the set of vertices and $W\in\mathbb{R}^{N\times N}$ is the weighted adjacency matrix. The entry $W_{ij}$ denotes the weight on the edge between vertices $x_i,x_j,~i\ne j$, which quantifies the similarity between the feature vectors $x_i$ and $x_j$. In general, the weight matrix is defined by
\begin{equation}\label{eq: gen_weights}
    W_{ij} = f\left(\frac{d(x_i,x_j)^2}{\sigma_i\sigma_j}\right),
\end{equation} 
where $d(x_i,x_j)$ is the distance between feature vectors $x_i$ and $x_j$, $f(\cdot)$ is the kernel function and $\sigma_i,\sigma_j$ are normalization constants corresponding to $x_i, x_j$, respectively. The choice of the distance function $d$ in this paper is the angular distance, i.e. 
\begin{equation}
    d(x,y) = \arccos\left(\frac{x^\top y}{\|x\|_2\|y\|_2}\right).\label{eq:angular_dist}
\end{equation}
The normalization constant $\sigma_i$ associated to node $i$ is chosen according to the distance to the $K^{th}$ nearest neighbor of $i$, i.e., $\sigma_i = \sqrt{d(x_i,x_{i_K})}$, where $x_{i_K}$ is the $K^{th}$ nearest neighbor to $x_i$. The kernel function $f(\cdot)$ is the Gaussian (exponential) kernel $f(x) = \text{exp}(-x)$. 

In order to improve computational efficiency, we only calculate the weight $W_{ij}$ between nearby pairs of nodes $x_i,x_j$. For each node $x_i$, we calculate edge weight $W_{ij}$ between $x_i$ and $x_j$ if and only if $x_j$ is among the $K$-nearest neighbors (KNN) of $x_i$ in the sense of the angular distance \eqref{eq:angular_dist}. We use an approximate nearest neighbor search algorithm \cite{arya1998optimal} to find the KNN of each node. This results in a KNN weight matrix $\bar{W}$ defined by
\begin{equation}
    \bar{W}_{ij} = 
    \begin{cases}
    &W_{ij}, \ j = i_1,i_2,\ldots,i_K,\\
    &0, \ \text{otherwise}.
    \end{cases}
\end{equation}
The parameter $K$ should be chosen to ensure that the corresponding graph $G$ is connected. We symmetrize the sparse weight matrix to obtain our final weight matrix by redefining $W_{ij} := (\bar{W}_{ij} + \bar{W}_{ji})/2$. Note that $W$ is sparse, symmetric and non-negative (i.e. $W_{ij} \ge 0$). In the following experiments of this paper, we choose the parameter $K = 50$ for the K-nearest neighbor search algorithm.

\subsection{Graph Learning}
With a graph $G=(X,W)$ constructed as described above, we now describe a graph-based approach for semi-supervised learning and present previous work in this field. We denote the ground truth one-hot encoding mapping by $\y^\dag: X_L\rightarrow \{e_1,e_2,\ldots,e_{n_c}\}$, $\y^\dag(x_j) = e_{y^\dag_j}$, where $y^\dag_j\in\{1,2,\ldots,n_c\}$ is the ground-truth label of the node $x_j$ and $e_k$ is the $k^\text{th}$ standard basis vector with all zeros except a $1$ at the $k^\text{th}$ entry. We have information on the ground truth labels for the subset $X_L\subset X$. The goal for the SSL task is to predict the labels of the \textit{unlabeled} nodes $x_i\in X\setminus X_L$. We are interested in solving for a classifier $\hat{\bfu}: X\rightarrow \mathbb{R}^{n_c}$ that reflects the probability of belonging to a certain class. In other words, $\hat{\bfu}(x_i)$ is a vector whose $j^\text{th}$ entry reflects the probability that $x_i$ belongs to class $j$. The graph-based SSL model that we consider obtains a classifier  by identifying $\hat{\bfu}$ with its matrix representation $\hat U_{i, j} = \hat{\bfu}(x_i)_j$ and solving the following optimization problem
\begin{equation}\label{eq:opt_gl}
\begin{split}
    \hat{U} &= \argmin_{U\in\mathbb{R}^{N\times n_c}}\ J_\ell(U, \y^\dag)=\argmin_{U\in\mathbb{R}^{N\times n_c}}\ \frac{1}{2}\langle U, LU\rangle_F + \sum_{j\in Z_0}\ell(\bfu(j),\y^\dag(j)),
\end{split}
\end{equation}
where $\langle \cdot,\cdot\rangle_F$ is the Frobenius inner product for matrices. In equation \eqref{eq:opt_gl}, $L = D - W$ is the unnormalized graph Laplacian matrix\cite{von2007tutorial}, where $D$ is the diagonal matrix with diagonal entries $D_{j,j} = \sum_{k\in Z} W_{jk}$ for $1\leq j\leq N$. 

The first term of \eqref{eq:opt_gl} optimizes over the smoothness of the classifier, whereas the second term imposes a penalty which ensures that the output of $\bfu$ at the labeled nodes stays close to the observed labelings $\y^\dag$. The loss function $\ell:\mathbb{R}^{n_c}\times \mathbb{R}^{n_c}\rightarrow\mathbb{R}$ measures the difference between the prediction $\bfu(x_j)$ and the ground-truth $\y^\dag(x_j)$ for $x_j\in X_L$. While there are several choices for the loss function, we simply apply a hard-constraint penalty 
\begin{equation}
    \ell_h(x,y) = \begin{cases}
    +\infty, \text{ if } x\ne y,\\
    0, \text{ if } x= y.
    \end{cases}
\end{equation}
This hard-constraint penalty function $\ell_h$ forces the minimizer $\hat{U}$ to be exactly the same as the ground-truth $\y^\dag$ on the observation set $X_L$. We refer to this SSL scheme as \textit{Laplace learning} \cite{zhu2003semi}. The predicted label of an unlabeled node $x_i\in X\setminus X_L$ of this graph SSL model is given by 
$$\hat{y}_i = \argmax_k\ \{\hat{\bfu}_k(x_i)|~k=1,2,\ldots,n_c\}.$$

\subsection{Active Learning}\label{sec:mb_al}

In domains with high labeling costs (ie time and/or money), the accuracy of a machine learning classifier is often constrained by the size of the training set. Active learning was developed to address this issue by providing machine learning algorithms with a way to quantify the value of obtaining labels from data. With a good determination of the labeling value, the machine learning algorithm seeks to maximize the amount of information it gets from each piece of data. This enables models trained using active learning to attain near-optimal performance with a relatively small training/labeled set \cite{settles_active_2012}. 

Active learning is an iterative process, whereby the model is continually improved through acquiring more labeled data. In each step, given a candidate set $\mathcal{C}$, active learning aims to select a query set $\mathcal{Q}\subset \mathcal{C}$ that the model considers ``most helpful'' to obtain labels for. Usually, the candidate set $\mathcal{C}$ is the set of current unlabeled nodes at each step. At the core of the active learning process is the acquisition function $\mathcal{A}:\mathcal{C}\rightarrow \mathbb{R}$, which quantifies the value of acquiring a label for each unlabeled node in the dataset. We consider four choices for the acquisition function, namely, uncertainty (UC)\cite{bertozzi2017uncertainty, miller2020efficient, qiao2019uncertainty}, Model-Change (MC)\cite{miller2020efficient, miller2021model}, Variance Optimality (VOpt)\cite{ji2012variance} and Model-Change Variance Optimality (MCVOpt) acquisition functions \cite{miller2022graph}. 

\begin{figure}
    \centering
    \includegraphics[width=0.9\textwidth]{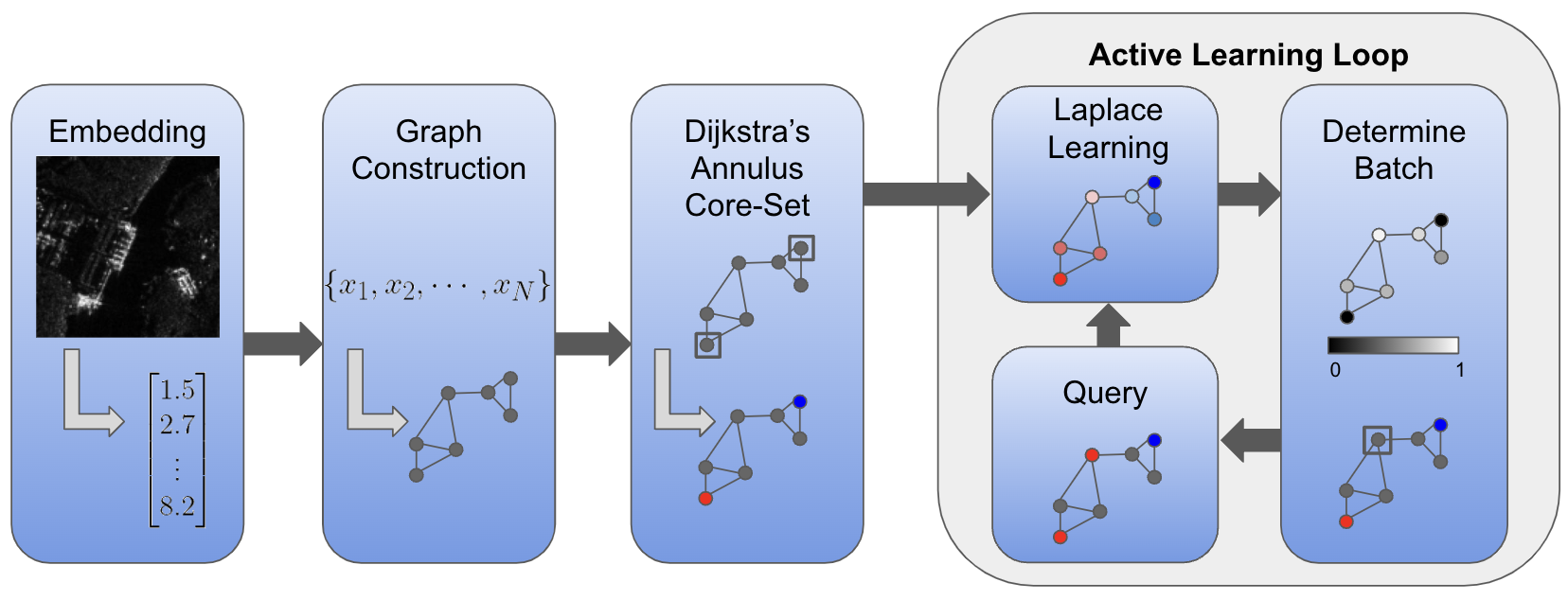}
    \caption{Flowchart of our batch active learning pipeline. First, images are embedded as feature vectors using either transfer learning or a CNNVAE. These feature vectors are then processed to create a weighted graph based on approximate $k$ nearest neighbors. Following this, the core-set is computed using DAC. Lastly, the graph is passed to the active learning loop. The active learning loop includes the classification of unlabeled nodes via Laplace learning, the evaluation of an acquisition function based on the classification and the selection of a query set according to our LocalMax method. After a batch is determined, the active learning loop queries a human to label the selected query set in each iteration manually. The solid red and blue nodes denote classifications of the data, while dark grey nodes denote unlabeled data. The SAR image shown above is from FUSAR-Ship\cite{hou2020fusar}.}
    \label{fig:al_pipeline}
\end{figure}

To recap, assume we want to classify a certain feature vector set $X = \{x_1,x_2,\ldots,x_N\}\subset\mathbb{R}^d$ through graph Laplace learning and active learning. The active learning process (illustrated in Figure~\ref{fig:al_pipeline}) starts with an initial labeled set $Y_0$. In the $k^\text{th}$ iteration, denote the current labeled set and current candidate set by $C_k = X\setminus Y_k$. We evaluate the acquisition function $\mathcal{A}_k$ on the candidate set $C_k$ then select the query set $\mathcal{Q}_k\subset C_k$. An oracle then labels the data in the query set $\mathcal{Q}_k$. Lastly, we update $Y_{k+1} = Y_k\cup \mathcal{Q}_k$ and $C_{k+1} = X\setminus Y_{k+1}$ before repeating the process. In practice, the oracle is a human who can provide ground-truth labels for datapoints in the query set.

\section{Core-Set Selection and Batch Active Learning}
\label{sec:balintro}

As introduced in Section~\ref{sec:mb_al}, active learning is an iterative process which selects a query set $\mathcal{Q}$ to obtain labels in each iteration. Sequential active learning selects one datapoint for each query (the case $\lvert\mathcal{Q}\rvert=1$)
while batch active learning selects multiple datapoints as the query set ( the case $\lvert\mathcal{Q}\rvert>1$). Typically, the human labeling time presents the largest bottleneck in the process. In sequential active learning, the human labeling process is limited by only being able to label one point at a time. It is advantageous for the active learning procedure to return a batch of points so that multiple humans/teams can work in parallel to label the data. Consider the example of labeling 200 points with a team of 10 people. In the sequential case $|\mathcal{Q}|=1$, labeling requires 200 human queries. In the batch case with $|\mathcal{Q}|=10$, only 20 human queries are required and the humans can label the data in parallel. This reduces the total labeling time by an order of magnitude as compared to the sequential case.

Many methods rely on using the current predictions of the model to evaluate the uncertainty, variance, or other criteria to quantify the information gained by labeling a new data point. These methods rely on a key assumption: the model has enough information to make a determination of what data would help it learn best. This leads to two interesting requirements: constructing a good set of initial labels, called a \textit{core-set}, for the early stages of active learning and quantifying information shared by unlabeled points in a batch. It is important that we construct a good core-set so that the model can make good estimates of the acquisition function early in training when model accuracy is relatively low. The core-set construction can have a long-term impact on the performance of the active learning procedure and it is critical that the method adequately explore the data before exploiting knowledge with active learning. Fortunately, graph-based active learning methods perform well with relatively small amounts of labeled data\cite{miller2022graph}. 

The main problem with transitioning from the sequential case to the batch case is that sequential active learning methods fail to capture shared information between unlabeled data points. Applying batch active learning naively often leads to batches of very similar points. For example, the $k$ points with the highest acquisition value tend to be close to one another in the embedding space. This leads to a great amount of redundancy in the batch and the model learns as fast as in the sequential case, but with more labeling done at each step. Therefore, batch active learning methods must encourage a diverse selection of points which high acquisition values. Proper implementation of batch active learning must also take into account the time needed to acquire good batches. Optimizing $\mathcal{Q}$ with $|\mathcal{Q}| = B$ over all subsets of the unlabeled data is a combinatorial problem with complexity $O(N^B)$. This time complexity would drastically limit the size of the batches that could be produced, so heuristics should be used to efficiently maximize the acquisition value of batches.

\subsection{Core-Set Selection}
The goal of core-set selection is to sufficiently explore the data so that the active learning procedure can perform well. We propose Algorithm~\ref{alg:DijCoreset} for core-set selection as it generates a core-set which is nearly uniform in the dataset. For a given feature vector set $X$, we construct a graph $G=(X,W)$ according to Section~\ref{sec:graph_construction}. The algorithm iteratively selects nodes in $X$ to construct a core-set such that all points are a distance at least $r$ from each other but no more than distance $R$ from another point. At each step of the core-set selection process, assuming the currently selected node set is $Y$ (i.e. current labeled set), the algorithm creates an \textit{annular set} $C$ and a \textit{seen set} $S$:
$$C = \left(\cup_{x\in Y} B_R(x)\right)\setminus \cup_{x\in Y}B_r(x)\qquad
S = \cup_{x\in Y} B_r(x),$$
where 
$$B_r(x) = \{y\in V(G):\ d_G(x, y) < r\}$$
and $d_G(x, y)$ is the distance from $x$ to $y$ computed using Dijkstra's algorithm \cite{dijkstra1959note}. The annular set is the set of points that we may select from at each stage in the algorithm and the seen set is a set of points that the algorithm can no longer select from. 

\begin{figure}
\centering
\begin{subfigure}{.18\textwidth}
  \centering
  \includegraphics[scale=.3]{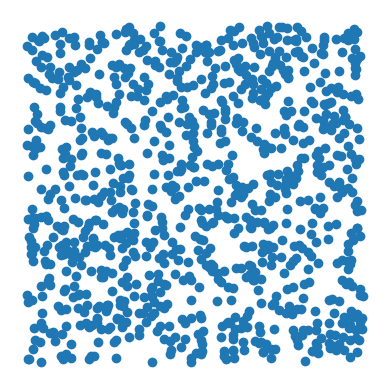}
  \caption{Iteration 0}
  \label{fig:DAC0}
\end{subfigure}
\begin{subfigure}{.18\textwidth}
  \centering
  \includegraphics[scale=.3]{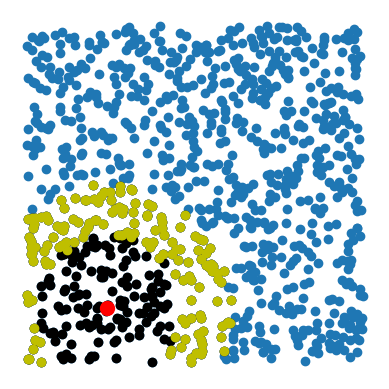}
  \caption{Iteration 1}
  \label{fig:DAC1}
\end{subfigure}
\begin{subfigure}{.18\textwidth}
  \centering
  \includegraphics[scale=.3]{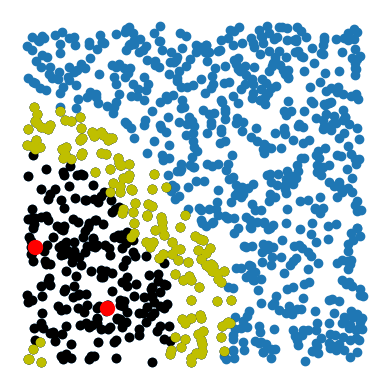}
  \caption{Iteration 2}
  \label{fig:DAC2}
\end{subfigure}
\begin{subfigure}{.18\textwidth}
  \centering
  \includegraphics[scale=.3]{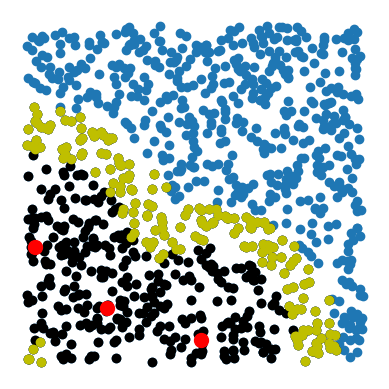}
  \caption{Iteration 3}
  \label{fig:DAC3}
\end{subfigure}
$\cdots$
\begin{subfigure}{.18\textwidth}
  \centering
  \includegraphics[scale=.3]{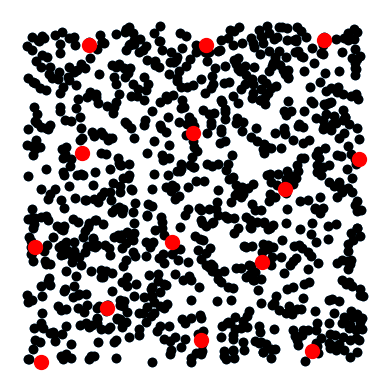}
  \caption{Iteration 14}
  \label{fig:DAC14}
\end{subfigure}
    \caption{An example of the sampling process of the DAC algorithm with an outer density radius of $0.3$. The dataset is generated by sampling uniformly at random in the unit square. The blue, black and gold points denote the unseen points, the seen points and the points in the annular set, respectively.  In iteration 0, the annular set is empty and the unseen set isn't empty. This means the algorithm picks a point at random from the unseen points to add to the core-set. In subsequent iterations, the algorithm picks a point at random from the annular set. It then updates the annular region as described in Algorithm~\ref{alg:DijCoreset}. This process terminates at iteration 14 when the entire dataset is the seen set. The set of red points in panel (e) is the output DAC core-set, which is nearly uniformly distributed in the whole dataset.}
    \label{fig:DijCoresetExample}
\end{figure}

Algorithm~\ref{alg:DijCoreset} then randomly selects $x\in C$ and updates $Y, S, C$. Repeating this process gives a somewhat uniform covering of the data. Note that it may not always be possible to select $x\in C$ before $S = X$. In this case, the algorithm randomly jumps to another data point outside of the seen set. This can occur if $R$ is too small or if there are well-separated clusters in the data. The output of Algorithm~\ref{alg:DijCoreset} is the core-set $Y$, which serves as the initial labeled set in the active learning process.  An example of the algorithm on a simple dataset is shown in Figure~\ref{fig:DijCoresetExample}.

\begin{algorithm}
\caption{Dijkstras Annulus Core-Set (DAC)}\label{alg:DijCoreset}
\begin{algorithmic}
\State Given: Graph $G = (X,W)$, initial labeled set $Y$, inner radius $r$ and outer radius $R$
\State Initialize candidate set $C = \emptyset$ and already seen set $S$
\For{ each $x\in Y$} \Comment{Compute candidate set from already labeled set}
    \State Compute $B_r(x), B_{R}(x)$
    \State $S\gets S\cup B_r(x)$
    \State $C\gets (C\cup B_{R}(x))\setminus B_r(x)$
\EndFor

\While{$S\neq X$} 
    \If{$C = \emptyset$}
        \State pick $x\in X\setminus S$ uniformly at random
    \ElsIf{$C\neq \emptyset$}
        \State pick $x\in C$ uniformly at random
    \EndIf
    \State Compute $B_r(x)$, $B_R(x)$
    \State $Y \gets Y\cup\{x\}$
    \State $S \gets S\cup B_r(x)$
    \State $C\gets (C\cup B_{R}(x))\setminus B_r(x)$
\EndWhile
\State Return $Y$
\end{algorithmic}
\end{algorithm}

Note that this algorithm is also used with $r, R$ adaptively determined by the density of the data around a point. For example, the user could specify that $r$ be picked so that $5\%$ of the data lies in $B_R(x)$. Using the density radius leads to greater exploration in high density regions and less exploration in low density regions. This causes the core-set to focus more on where the majority of the data lies. Additionally, the density based covering is independent of the average distances between data points, which reduces the need for parameter tuning. We further reduce parameter tuning by setting $r = R/2$.

\subsection{LocalMax}
\label{sec:local_max}

\begin{figure}
    \centering
\begin{subfigure}{.28\textwidth}
  \centering
  \includegraphics[scale=.45]{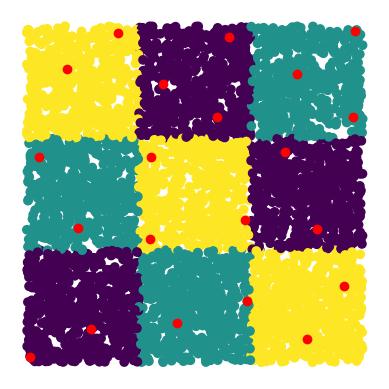}
  \caption{Ground Truth}
  \label{fig:LM_checkerboard}
\end{subfigure}
\begin{subfigure}{.28\textwidth}
  \centering
  \includegraphics[scale=.45]{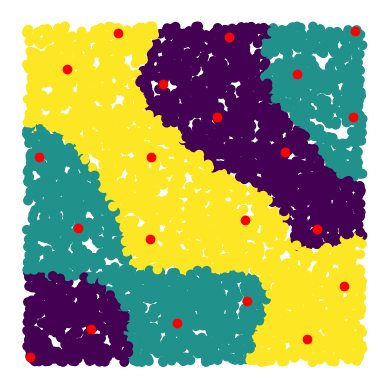}
  \caption{Classifier}
  \label{fig:LM_classifier}
\end{subfigure}
\begin{subfigure}{.34\textwidth}
  \centering
  \includegraphics[scale=.45]{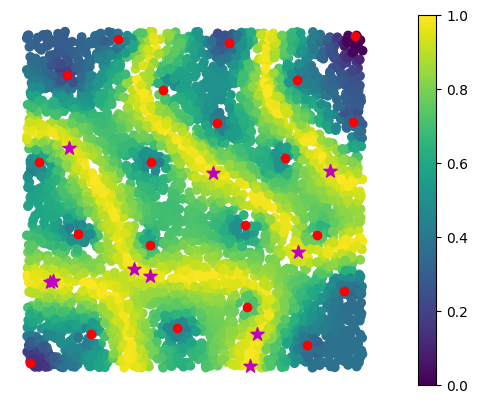}
  \caption{Acquisition Values}
  \label{fig:LM_acq_fun}
\end{subfigure}
    \caption{An example of DAC and LocalMax on the checkerboard dataset. In all panels, red points denote the labeled core-set generated by DAC. Panel (a) shows the ground truth classification. Panel (b) shows the classification results of Laplace learning based on the labeled core-set. Panel (c) shows the heatmap of the uncertainty acquisition function evaluated on the dataset. For the uncertainty acquisition function, high acquisition values concentrate near the decision boundary. In panel (c), the purple stars denote points in the query set returned by LocalMax with a batch size of $10$. }
    \label{fig:LM_example}
\end{figure}

We propose a novel batch active learning approach, named LocalMax. Based on a feature vector set $X$ and the corresponding similarity graph $G$, LocalMax selects a query set of multiple nodes that satisfy the local maximum condition (Definition~\ref{def:localmax}) on the graph $G$ from the candidate set. Informally, a node is a local maximum of a function on the nodes if and only if its function value is at least that of its neighbors. 

\begin{definition}[\textbf{Local Max of a Graph Node Function}]\label{def:localmax}
    Consider a KNN-generated graph $G = (X,W)$, where $X$ is the set of nodes and $W$ is the edge weight matrix. For a graph node function $\mathcal{A}: X\rightarrow \mathbb{R}$, $x_i\in X$ is a \textbf{local maximum node} if and only if for any $x_j$ adjacent to $x_i$, $\mathcal{A}(x_i)\ge  \mathcal{A}(x_j)$. Equivalently, $x_i\in X$ is a local maximum if and only if:
    \begin{equation}
        \mathcal{A}(x_i)\ge \mathcal{A}(x_j),~\forall j \text{ s.t. }W_{ij}>0.
    \end{equation}
\end{definition}

Assuming a batch size $B$, at iteration $k$ LocalMax selects the query set $\mathcal{Q}_k$ as the top-$B$ local maximums in the candidate set $\mathcal{C}_k$ (as detailed in Algorithm~\ref{alg:localmax}). LocalMax benefits from many useful properties including simplicity, efficiency and its grounding in well studied sequential acquisition functions. Building this acquisition function on sequential active learning allows us to borrow properties from the sequential acquisition functions. Controlling for local maxes enforces a minimum pairwise distance between points in the query set, which counteracts the redundancy seen in naively optimizing sequential acquisition functions. 

\begin{algorithm}
\caption{LocalMax Batch Active Learning}\label{alg:localmax}
\begin{algorithmic}[1]
\Require A KNN graph $G=(X,W)$, $X$ is indexed by $Z$. A labeled index set $Z_0$. An acquisition function $\mathcal{A}:Z-Z_0\rightarrow \mathbb{R}^+$. Batch size $B$.
\Ensure The query set $\mathcal{Q}$.
\State Extend the domain of $\mathcal{A}$ to $Z$ by defining $\mathcal{A}(j)=0,~\forall j\in Z_0$
\State $S \leftarrow Z-Z_0$; $\mathcal{Q}=\emptyset$
\While{$S\ne \emptyset$ and $\lvert \mathcal{Q}\rvert<B$}
\State $k\leftarrow \argmax_{k\in S}\ \mathcal{A}(k)$
\State $N(k) = \{i\in Z: W_{jk}>0\}$
\If{$\mathcal{A}(k)\ge \mathcal{A}(i),~\forall i\in N(k)$}
\State $Q\leftarrow Q\cup \{k\}$
\EndIf
\State $S\leftarrow S-N(k)$
\EndWhile
\end{algorithmic}
\end{algorithm}

LocalMax also maintains good computational complexity. The computational complexity of Algorithm~\ref{alg:localmax} is $O(KN)$ where $N$ is the number of nodes in the graph and $K$ is the KNN parameter used when generating the graph. In practice, $K$ is much smaller than $N$, so the computational complexity is $O(N)$. Let $\mathcal{G}(N)$ denote the model fitting time for Laplace learning on a graph $G$ with $N$ nodes and let $\mathcal{H}$ denote the human labeling time for each node. In the case that the weight matrix $W$ is sparse and the graph Laplacian matrix $L$ is well-conditioned, we have $\mathcal{G}(N) = O(N)$. The human labeling time, $\mathcal{H}$, is typically much greater than $O(N)$ since labeling SAR data can take significantly more time than is required by the rest of the active learning pipeline. Since human labeling can be processed in parallel, fewer batches are required to label the same amount of data. Consider the active learning process that samples in total $m$ nodes to obtain ground-truth labels. The time complexities of sequential active learning and LocalMax batch active learning with batch size $B$ are:
\begin{align}
    \text{Sequential Active Learning:}&~~~ m\times O(\mathcal{G}(N)+\mathcal{H}) && = O(m\mathcal{H}),\\
    \text{LocalMax Batch Active Learning:}&~~~ m/B\times O(\mathcal{G}(N)+O(N)+\mathcal{H}) && = O(m\mathcal{H}/B).
\end{align}
Since $\mathcal{G}(N)=O(N)$ and $\mathcal{H}\gg O(N)$, LocalMax provides a $B$ times speed up based on sequential active learning which is observed both theoretically and in practice as shown in Table~\ref{table:timeandacc}.

\section{EXPERIMENTS AND RESULTS}
\label{sec:expintro}
We now present the results of our active learning experiments on the MSTAR, OpenSARShip, and FUSAR-Ship datasets. We first test the accuracy and efficiency of our methods as compared to other batch active learning methods and sequential active learning. We then test the accuracy of our methods with different embedding techniques on each dataset. Lastly, we look at the impact of data augmentation and the choice of neural network architecture in transfer learning. It is important to note that our methods use less data than state-of-the-art methods. LocalMax surpasses state-of-the-art accuracy on OpenSARShip, and FUSAR-Ship with $35\%$ and $68\%$ of the data labeled, respectively. The previous state of the art utilized $44\%$ and $70\%$ of the data for OpenSARShip and FUSAR-Ship, respectively\cite{9445223}. 

In our transfer learning experiments, we use PyTorch CNNs pretrained on the ImageNet dataset to perform image classification on SAR datasets. Unless otherwise stated, we use AlexNet for OpenSARShip and ShuffleNet for FUSAR-Ship since preliminary experiments suggested that they would achieve the best performance among the neural networks tested. The transfer learning results for MSTAR were generally poor and we only present the results with transfer learning from ResNet. These choices are later examined in our comparison between different neural network architectures. Our experiments use both zero-shot and fine-tuned transfer learning. In all the embeddings mentioned, the data is first transformed using the following PyTorch data transformations in order: Resize, CenterCrop, RandomRotation, GaussianBlur, and ColorJitter. Also, LocalMax may not always find batches of the specified size (15), so it selects up to 15 points in a batch. As evidenced by the later efficiency improvements, we see that this occurs infrequently. 

Table~\ref{table:parameters} contains all the parameters used in our experiments. The experiment time measures the time taken to complete the entire active learning process after the core-set selection, including batch selection and model fitting. The time calculation neglects the human labeling time, so the performance enhancements seen in practice will be much larger. Accuracy is measured as the percent of correct predictions by the model in the unlabeled dataset. The source code to reproduce all the results is available\cite{graphlearning, calder2020poisson, scikit-learn}. All experiments were performed in Google Colab with high RAM.

\begin{table}[!ht]
\centering
\subfloat[][General Parameters]{
\centering
\begin{tabular}{|c|c|}
\hline
\rule{0pt}{12pt} Parameter & Value\\
\hline
\rule{0pt}{12pt} Batch size & 15\\
\hline
\rule{0pt}{12pt} Transfer learning data & 5\%\\
\hline
\rule{0pt}{12pt}Sequential Acquisition Function & Uncertainty\\
\hline
\end{tabular} 
\label{table:parameters_general}
}
\qquad
\subfloat[][Dataset Specific Paramters]{
\centering
\begin{tabular}{|c|c|c|}
\hline
\rule{0pt}{12pt} & Final Labels (\%) & TL Architecture \\
\hline
\rule{0pt}{12pt} MSTAR & 15\% & ResNet\\
\hline
\rule{0pt}{12pt} OpenSARShip & 35\% & AlexNet\\
\hline
\rule{0pt}{12pt} FUSAR-Ship & 68\% & ShuffleNet\\
\hline
\end{tabular} 
\label{table:parameters_dataset}
}
\vspace{0.3em}
\caption{Tables of parameters used in our experiments. All experiments use these parameters unless otherwise stated. In Table~\ref{table:parameters_general}, "transfer learning data" refers to the amount of data used in fine-tuned transfer learning. This data is sampled uniformly at random and is then used as part of the core-set before performing DAC. In Table~\ref{table:parameters_dataset}, "final labels" refers to the size of the labeled dataset as a percent of the total dataset size at the end of the active learning process. Also, "TL architecture" refers to the pretrained PyTorch neural network used for transfer learning on each dataset. }
\label{table:parameters}
\end{table}

\subsection{Accuracy And Efficiency}
\label{sec:acc_eff}
As seen in Table~\ref{table:timeandacc}, LocalMax generally outperforms the other batch active learning methods. Among the batch active learning methods tested, LocalMax attained the highest accuracy with comparable efficiency to the other methods. The time efficiency of LocalMax is slightly worse than random sampling, but this is due to the fact that random sampling is a very naive approach which achieves much lower accuracy than LocalMax in each dataset. The TopMax method has comparable time and accuracy to LocalMax, but in each test the accuracy is lower. This is likely due to the fact that TopMax does not enforce separation of the data in a batch and may select points with lots of shared information. Acq Sample also performs worse than LocalMax by a noticeable amount. 

We can see from Table~\ref{table:timeandacc} that LocalMax has comparable accuracy to sequential active learning, deviating by at most $0.64\%$ on each dataset. Additionally, LocalMax uses an order of magnitude less computational time than sequential active learning. The discrepency between theoretical efficiency analysis and the experimental time efficiency is due to the fact that human querying time is negligible in these experiments. The code immediately provides ground truth labels when queried, so most of the time in these experiments is observed in the model fit time. In practice, these time differences will better match theoretical predictions where the human query time is the bottleneck. These experiments attest to the efficiency and accuracy of LocalMax relative to other active learning methods. 

\begin{table}[!t]
\begin{center}
\begin{tabular}{|c|c|c|c|c|c|c|}
\hline
\multicolumn{2}{|c|}{ }& \textit{LocalMax} & \textit{Random} & \textit{TopMax} & \textit{Acq\_sample} & \textit{Sequential}\\
\hline
\multirow{3}{*}{\textit{\textbf{Time Consumption}}} 
&\textit{MSTAR} & 26.70s & \textbf{24.17s} & 25.96s & 26.71s & 338.11s\\
\cline{2-7}
&\textit{OpenSARShip} & 5.33s & \textbf{4.68s} & 4.86s & 4.86s &  47.43s\\
\cline{2-7}
&\textit{FUSAR-Ship} & 26.80s & \textbf{21.21s} & 26.08s & 21.55s & 322.99s \\
\hline
\multirow{3}{*}{\textit{\textbf{Accuracy}}} 
&\textit{MSTAR} & 99.69\% & 92.66\% & 99.69\% & 96.27\% & \textbf{99.93}\%\\
\cline{2-7}
&\textit{OpenSARShip}  & 81.25\% & 71.60\% & 80.64\% & 73.34\% &  \textbf{81.65\%}\\
\cline{2-7}
&\textit{FUSAR-Ship} & \textbf{89.83\%} & 68.73\% & 85.59\% & 72.72\% & 89.19\% \\
\hline
\end{tabular}
\end{center}
\caption{Time consumption and accuracy comparison among different active learning methods. This experiment uses a CNNVAE embedding for MSTAR and zero-shot transfer learning for OpenSARShip and FUSAR-Ship. Additionally, the parameters for all the methods are listed in Table~\ref{table:parameters}. LocalMax is the batch sampling method introduced in Section~\ref{sec:local_max}. Random is a batch active learning method which randomly chooses a new batch with the desired size. TopMax is a batch active learning method which chooses the $n$ points with highest acquisition values. The acq\_sample method assigns each point with a probability to be picked proportional to the acquisition value, and randomly samples $n$ points as a batch. All batch active learning methods have comparable efficiency and are 9 to 15 times faster than the sequential case. The local max method always achieved higher accuracy than other batch active learning methods and is comparable to the accuracy of sequential active learning. 
}
\label{table:timeandacc}
\end{table}

\begin{figure}[!ht]
\begin{subfigure}{.5\textwidth}
  \centering
  \includegraphics[width=.8\linewidth]{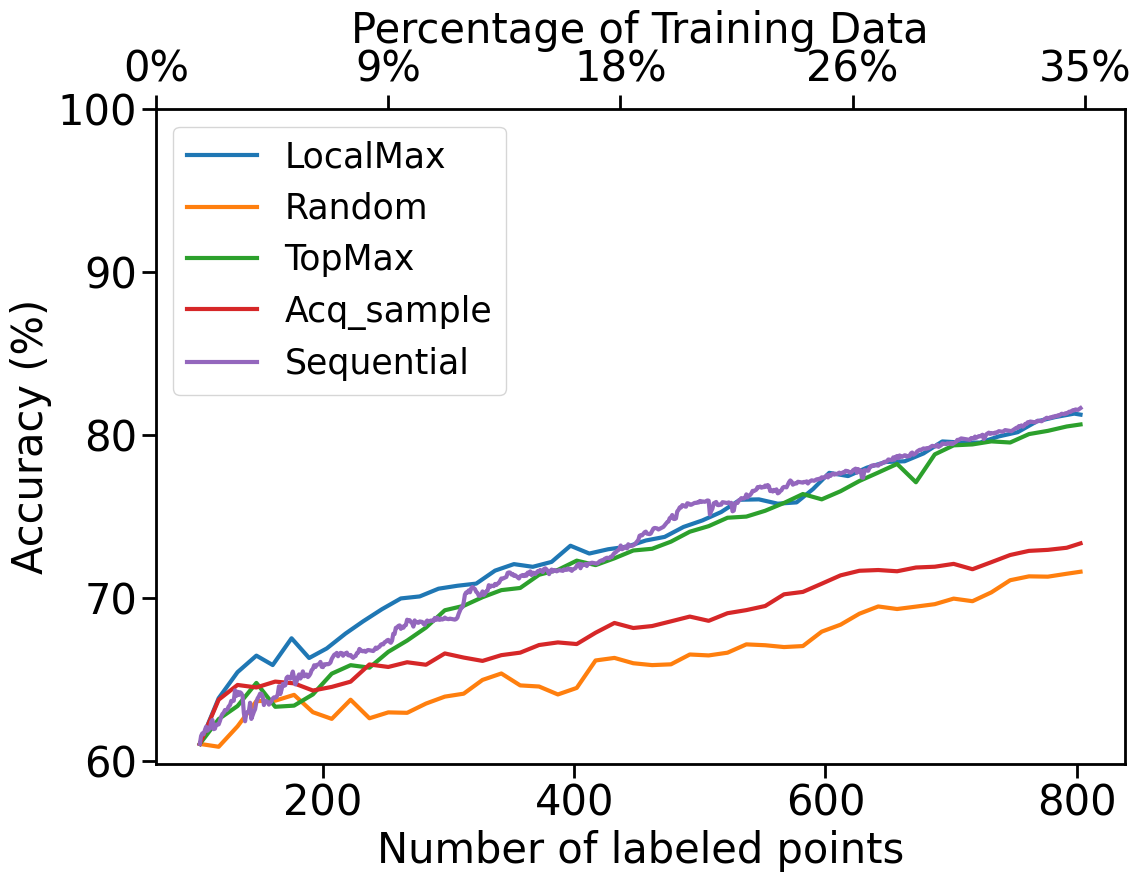}
  \caption{OpenSARShip}
  \label{fig:opensartime}
\end{subfigure}
\begin{subfigure}{.5\textwidth}
  \centering
  \includegraphics[width=.8\linewidth]{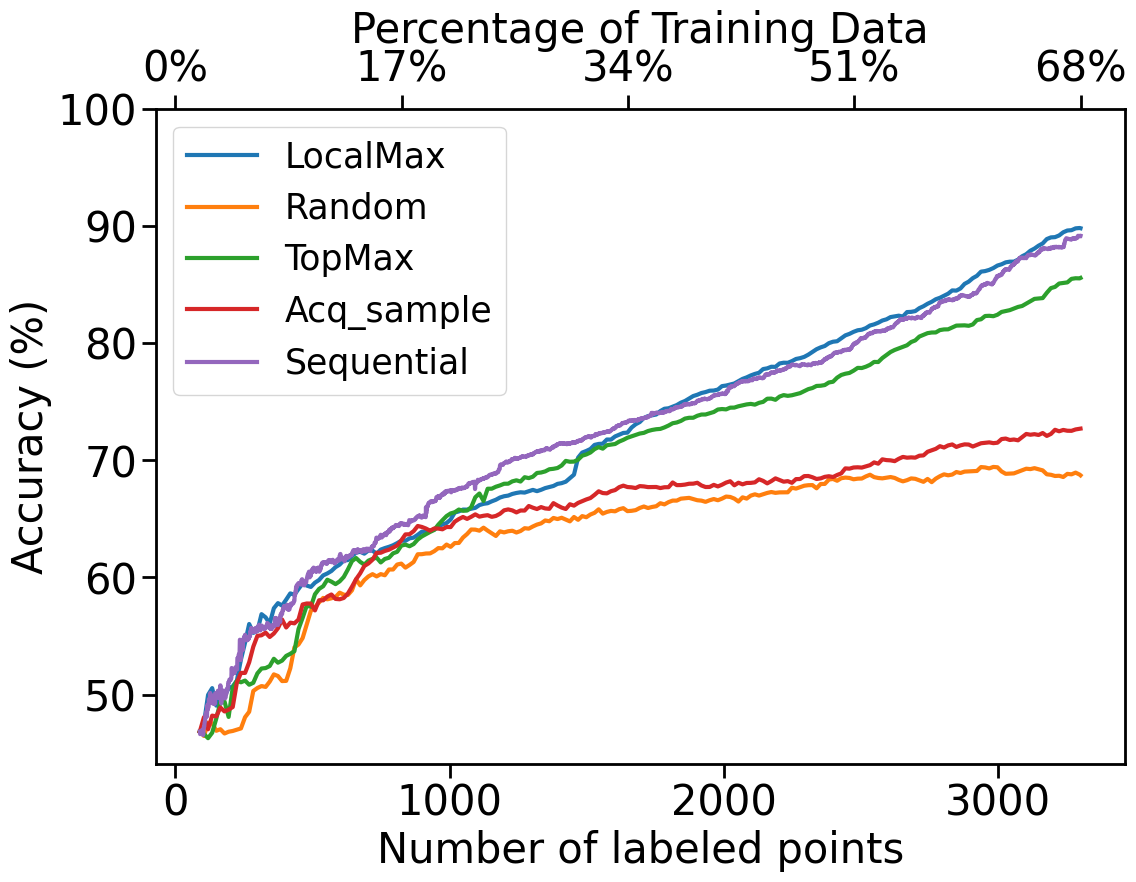}
  \caption{FUSAR-Ship}
  \label{fig:vaeresult}
\end{subfigure}
\vspace{0.5em}
\caption{Plots of accuracy v.s. the number of labeled points for five different active learning methods. Details about these active learning methods are shown in the caption of Table~\ref{table:timeandacc}. Panel (a) and (b) contain the results for the OpenSARShip and FUSAR-Ship datasets, respectively. In each panel, our LocalMax method (blue curve) and the sequential active learning (purple curve) are almost identical and are the best performing methods. According to Table~\ref{table:timeandacc}, LocalMax is much more efficient, proportional to the batch size. 
}
\label{fig:acc_vs_data}
\end{figure}

More detailed plots of the accuracy can be seen in Figure~\ref{fig:acc_vs_data}. These plots show accuracy as a function of the size of the labeled set. In each case, we see a large jump in accuracy with few labels, which is characteristic of the active learning process. After this point, the accuracy appears to grow linearly with the amount of labeled data. Again, we notice that LocalMax and sequential active learning tend to perform best on both datasets. We can also see that random sampling fails to capture the importance of labeling certain data as its accuracy is much lower throughout the active learning process. This shows that LocalMax is improving the model in a more substantial way than by just increasing the size of the labeled dataset. 

\begin{figure}[!ht]
\begin{subfigure}{.33\textwidth}
  \centering
  \includegraphics[width=1\linewidth]{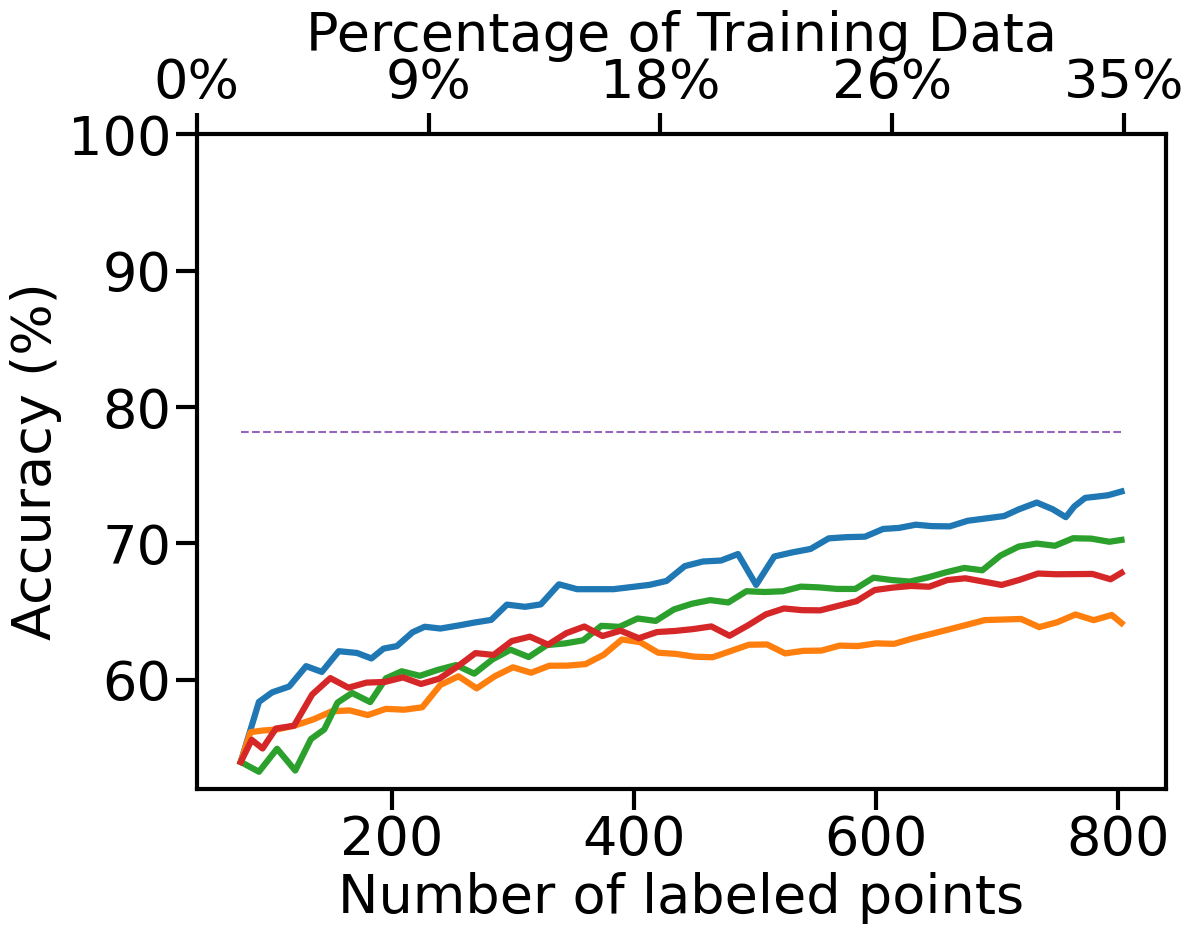}
  \caption{OpenSARShip CNNVAE Embedding}
  \label{fig:sfig1}
\end{subfigure}
\begin{subfigure}{.33\textwidth}
  \centering
  \includegraphics[width=1\linewidth]{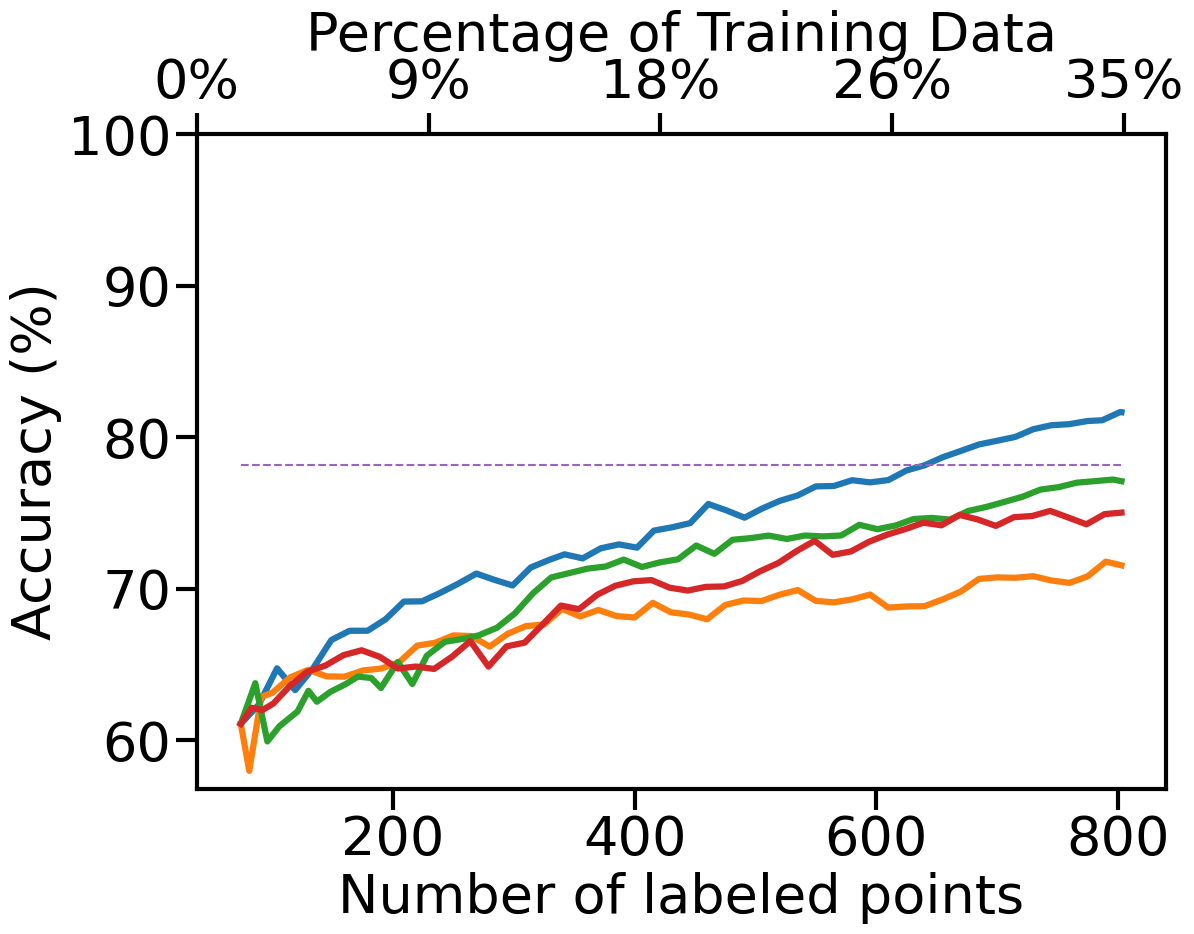}
  \caption{OpenSARShip Zero-shot Embedding}
  \label{fig:sfig2}
\end{subfigure}
\begin{subfigure}{.33\textwidth}
  \centering
  \includegraphics[width=1\linewidth]{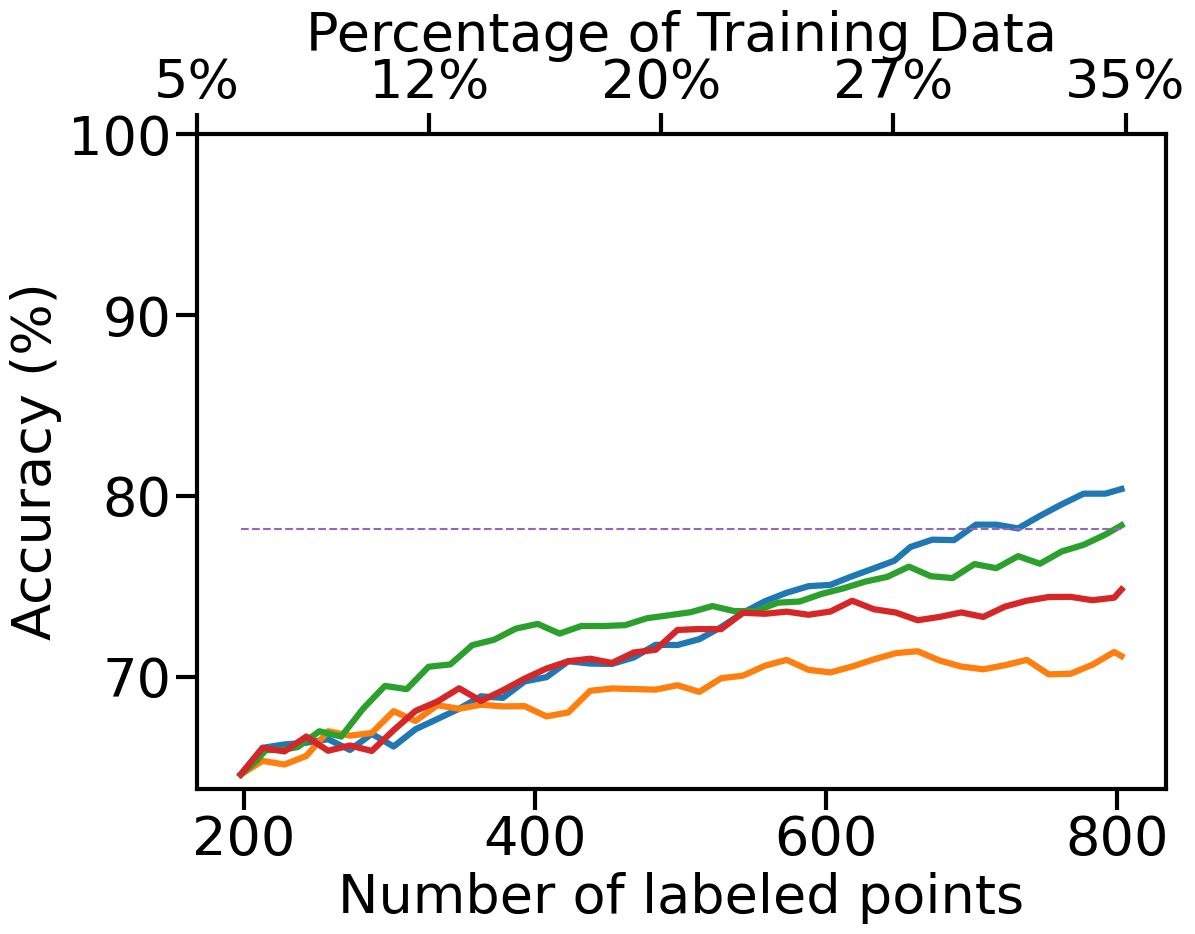}
  \caption{OpenSARShip Fine-tuned Embedding}
  \label{fig:sfig3}
\end{subfigure}\\
\vskip3ex
\begin{subfigure}{.33\textwidth}
  \centering
  \includegraphics[width=1\linewidth]{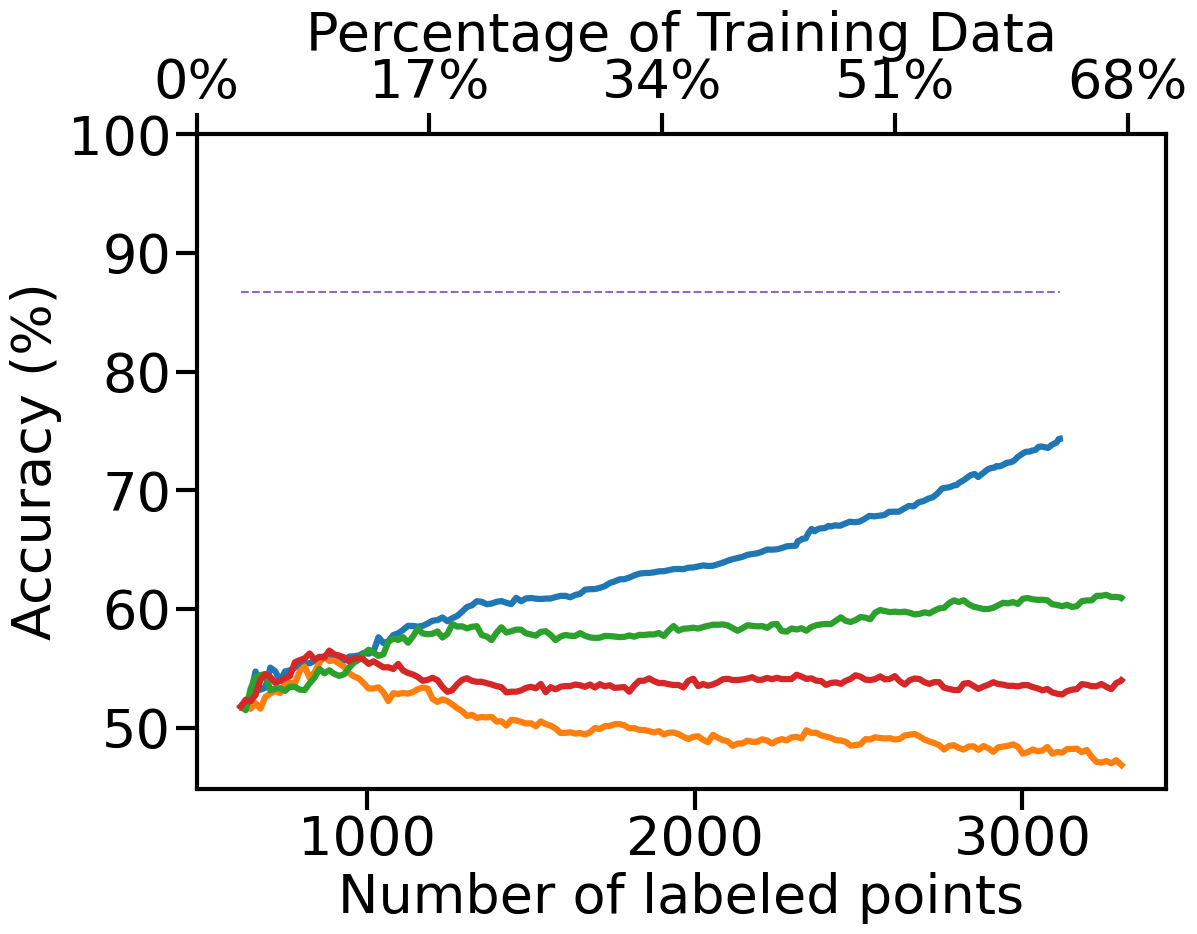}
  \caption{FUSAR-Ship CNNVAE Embedding}
  \label{fig:sfig4}
\end{subfigure}
\begin{subfigure}{.33\textwidth}
  \centering
  \includegraphics[width=1\linewidth]{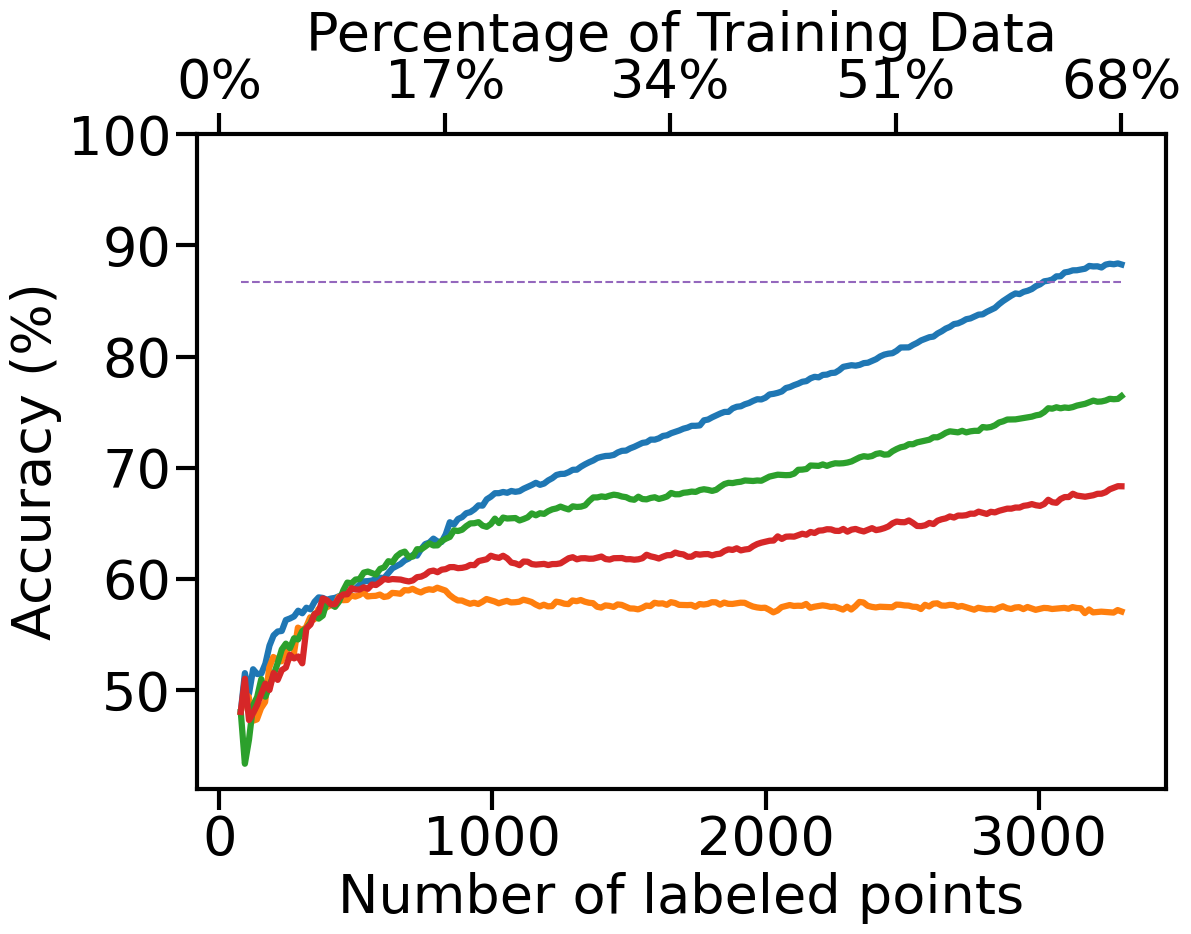}
  \caption{FUSAR-Ship Zero-shot Embedding}
  \label{fig:sfig5}
\end{subfigure}
\begin{subfigure}{.33\textwidth}
  \centering
  \includegraphics[width=1\linewidth]{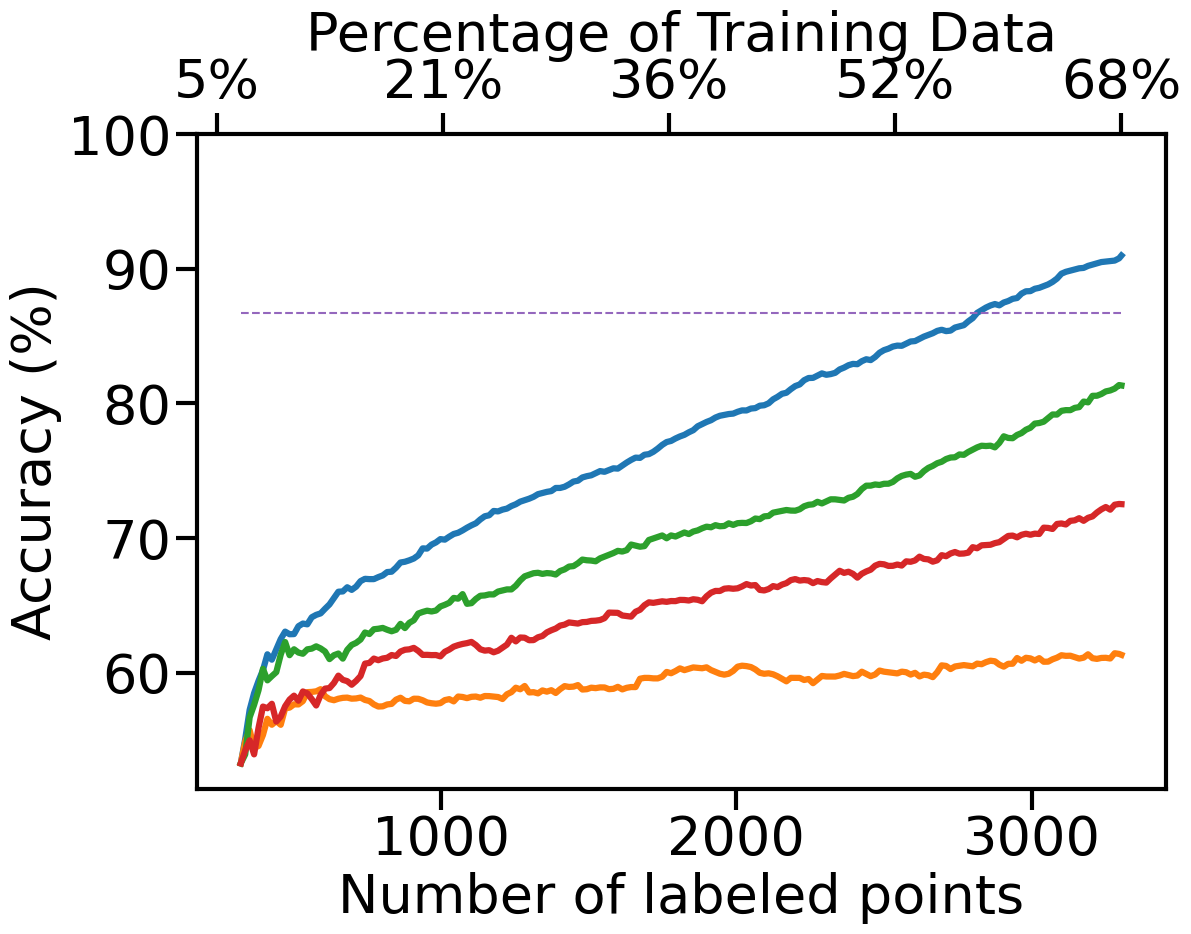}
  \caption{FUSAR-Ship Fine-tuned Embedding}
  \label{fig:sfig6}
\end{subfigure}\\
\vskip3ex
\begin{subfigure}{.33\textwidth}
  \centering
  \includegraphics[width=1\linewidth]{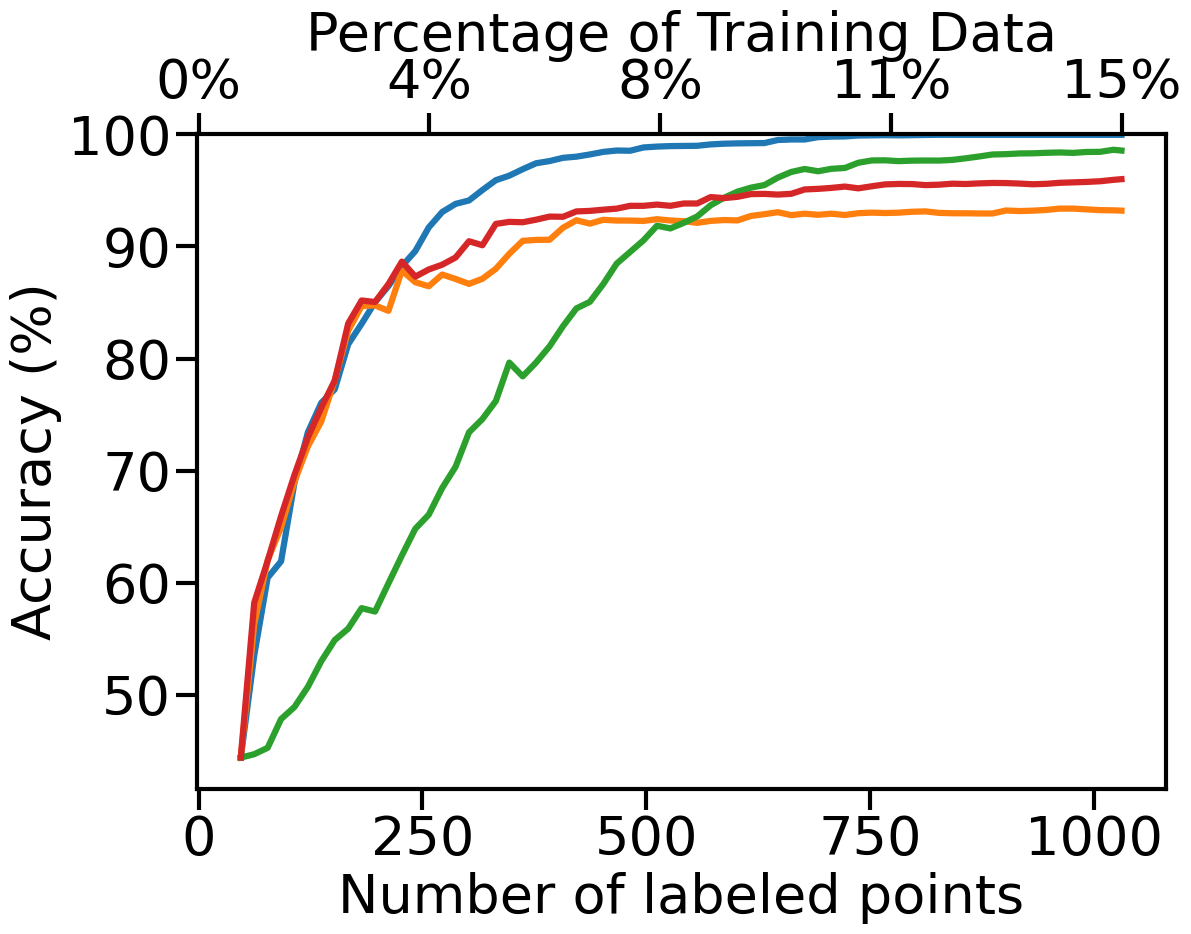}
  \caption{MSTAR CNNVAE Embedding}
  \label{fig:sfig7}
\end{subfigure}
\begin{subfigure}{.33\textwidth}
  \centering
  \includegraphics[width=1\linewidth]{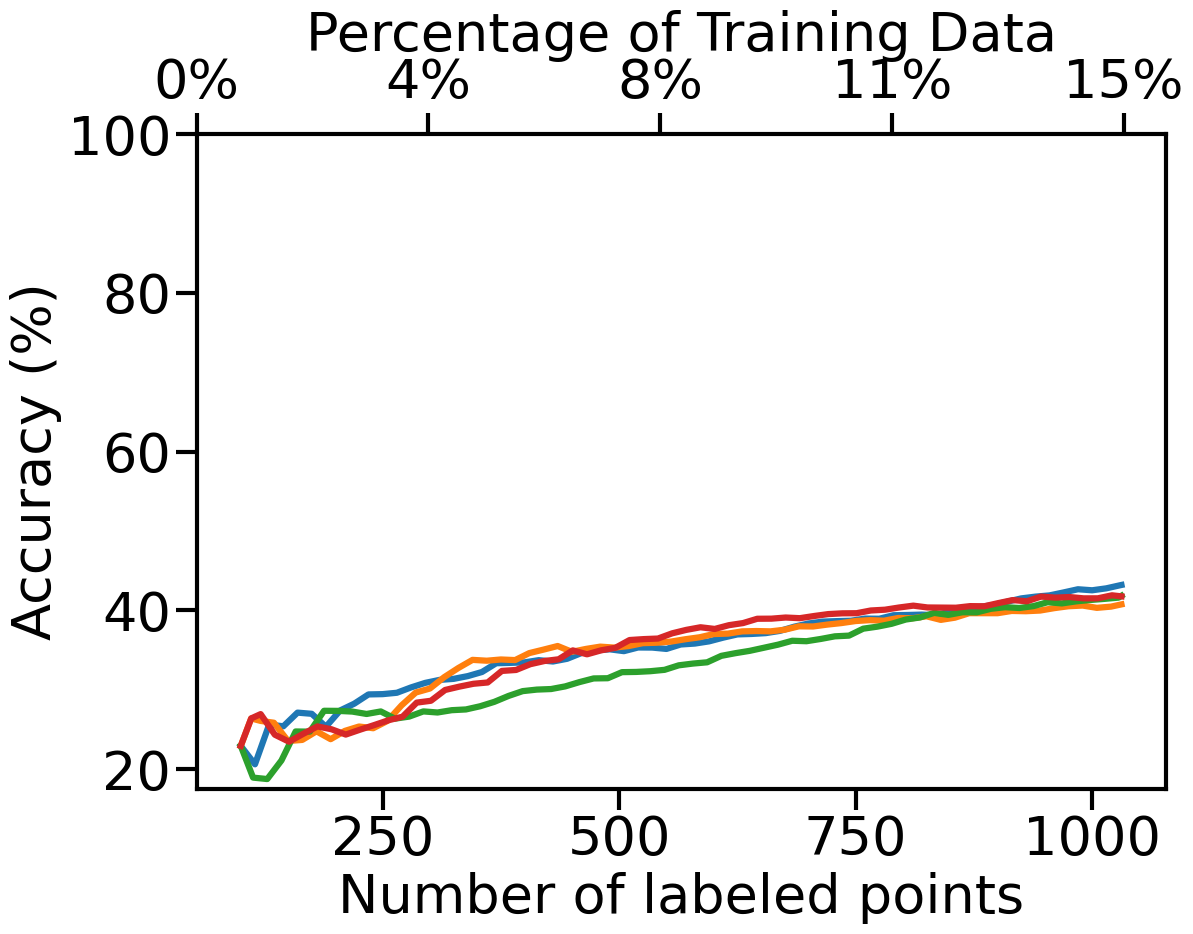}
  \caption{MSTAR Zero-shot Embedding}
  \label{fig:sfig8}
\end{subfigure}
\begin{subfigure}{.33\textwidth}
  \centering
  \includegraphics[width=1\linewidth]{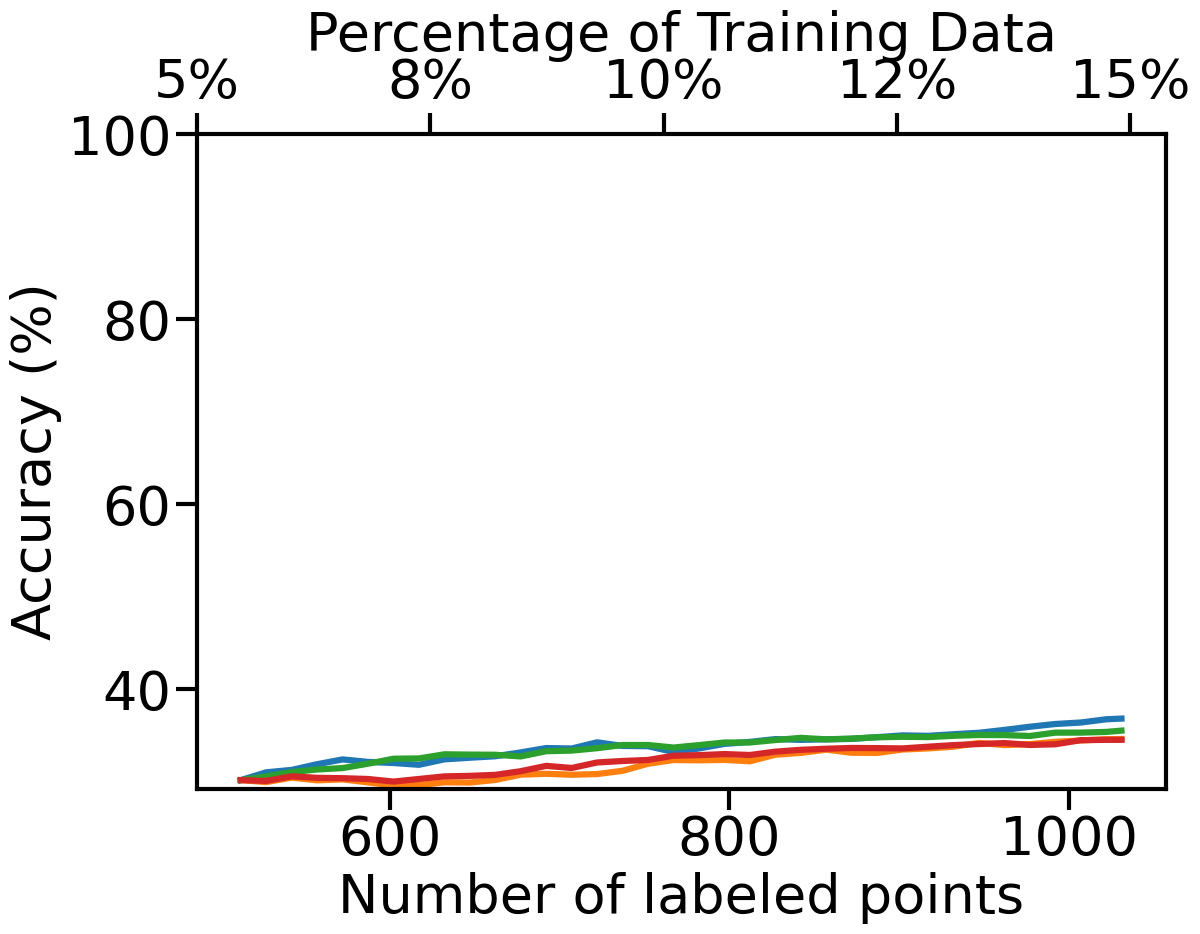}
  \caption{MSTAR Fine-tuned Embedding}
  \label{fig:sfig9}
\end{subfigure}
\begin{subfigure}{\textwidth}
    \centering
    \includegraphics[width=.5\linewidth]{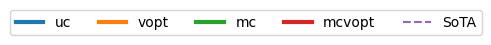}
\end{subfigure}
\vspace{0.5em}
\caption{Plots of accuracy v.s. number of labeled points for each embedding and dataset. In each panel, we show four curves generated by LocalMax using different active learning acquisition functions, UC, VOpt, MC and MCVOpt (Section~\ref{sec:mb_al}), together with the state-of-the-art, CNN-based method (denoted SoTA \cite{9445223}).
There are nine panels in total - the Three rows from top to bottom correspond to the OpenSARShip, FUSAR-Ship and MSTAR datasets while three columns from left to right correspond to CNNVAE embedding, zero-shot transfer learning embedding and fine-tuned transfer learning embedding. The UC acquisition function has the best performance among all the acquisition functions tested. The parameters for these experiments are the same as those specified in Table~\ref{table:parameters}. 
}
\label{fig:transferlocalmax}
\end{figure}

Following the comparison between different batch and sequential active learning methods, we analyze the impact of the acquisition function on LocalMax. The results of these experiments are shown in Figure~\ref{fig:transferlocalmax}, where we see that the uncertainty acquisition function performs best in all experiments. This matches results from previous works on sequential active learning \cite{miller2022graph}. Figure~\ref{fig:transferlocalmax} also shows that the uncertainty based LocalMax beats state of the art in both transfer learning experiments on FUSAR-Ship and OpenSARShip.

\subsection{Sensitivity Analysis}
\label{sec:sensitivity}
We now look at the sensitivity of our results based on choices in the pipeline. We first look at the impacts of data augmentation and fine tuning on the final results. Table~\ref{table:accvariation} contains summary statistics for an experiment regarding the benefits to using data augmentation and fine tuning in the transfer learning portion of the pipeline. The results of this experiment are not conclusive between OpenSARShip and FUSAR-Ship. For OpenSARShip, we see that variance is consistently low for the embeddings and zero-shot transfer learning performed best. In contrast, the FUSAR-Ship results had notably higher accuracy and variance for fine-tuned transfer learning without data augmentation. In fact, adding data augmentation reduced variance for fine-tuned transfer learning on both datasets. It is also important to note that each of these experiments showed better accuracy than the previous state of the art. Additionally, zero-shot transfer learning may be the most practical method as it attains comparable performance to the other embeddings, does not require labels in the new dataset, and has no training time.

\begin{table}[!ht]
    \centering
    LocalMax Accuracy with Different Embeddings
    \vspace{0.3em}
    \begin{tabular}{|c|c|c|c|c|}
        \hline
        \rule{0pt}{12pt} & State of the Art & Zero-shot & Fine-tuned & Fine-tuned + Augmentation \\
        \hline
        \rule{0pt}{12pt} OpenSARShip & $78.15\%\pm 0.57\%$ & $81.02\%\pm 0.76\%$ & $79.66\%\pm 0.90\%$ & $80.00\%\pm 0.75\%$\\
        \hline
        \rule{0pt}{12pt} FUSAR-Ship & $86.69\%\pm 0.47\%$ & $88.57\%\pm 0.35\%$ & $91.54\%\pm 3.07\%$ & $89.06\%\pm 1.90\%$\\
        \hline
    \end{tabular}
    \caption{
    Sample statistics of accuracy after 20 experiments of the batch active learning pipeline with zero-shot transfer learning, fine-tuned transfer learning, and fine-tuned transfer learning with data augmentation (laTst column). 
    The number in each cell represents the mean $\pm$ one standard deviation across the 20 experiments. The zero-shot and fine-tuned embeddings are the same as mentioned in Section~\ref{sec:embedding}. The parameters in these experiments are the same as those specified in Table~\ref{table:parameters}. } 
    \label{table:accvariation}
\end{table}

Lastly, we study the impact of neural network architecture on model performance. Table~\ref{table:cnncomparison} shows the results of one run of LocalMax for different choices of neural network architectures. The range of model performance across architectures is $8.10\%$ for OpenSARShip and $4.82\%$ for FUSAR-Ship. Although this is somewhat large variation, this is a common issue encountered in deep supervised learning. It is important to note that the neural network architectures tested are standard neural networks and weren't designed for SAR data. It is possible that transfer learning from architectures designed for SAR data will experience less variance in the final accuracy\cite{inkawhich2021bridging, arnold2018blending}.

\begin{table}[!ht]
\centering
Zero-shot Embedding Active Learning Various CNN\\
\vspace{0.3em}
\begin{tabular}{|c|c|c|}
\hline
\rule{0pt}{12pt} & OpenSARShip & FUSAR-Ship \\
\hline
\rule{0pt}{12pt} AlexNet & \textbf{80.24\%} & 85.14\%\\
\hline
\rule{0pt}{12pt} ResNet18 & 72.14\% & 87.39\%\\
\hline
\rule{0pt}{12pt} ShuffleNet & 72.34\% & 88.22\%\\
\hline
\rule{0pt}{12pt} DenseNet & 75.69\% & \textbf{89.96\%}\\
\hline
\rule{0pt}{12pt} GoogLeNet & 73.28\% & 86.74\%\\
\hline
\rule{0pt}{12pt} MobileNet V2 & 74.15\% & 86.68\%\\
\hline
\rule{0pt}{12pt} ResNeXt & 76.29\% & 88.29\%\\
\hline
\rule{0pt}{12pt} Wide ResNet & 73.14\% & 85.52\%\\
\hline
\end{tabular} 
\vspace{0.3em}
\caption{Accuracy values of one run of LocalMax for different choices of neural networks. Each experiment uses zero-shot transfer learning (no fine-tuning) and the parameter values specified in Table~\ref{table:parameters}. The highest accuracy value in each column is bolded. As shown in the table, the range of model performance across architectures is $8.10\%$ and $4.82\%$ for OpenSARShip and FUSAR-Ship, respectively.}
\label{table:cnncomparison}
\end{table}

\section{DISCUSSION}
\label{sec:discussion}
Improving our understanding of SAR data and the methods used for ATR is critical for designing new ATR algorithms. This paper seeks to understand the latter but gives rise to some surprising results the about SAR datasets tested. In particular, it is surprising to find that neural networks trained on ImageNet could perform well when applied to OpenSARShip and FUSAR-Ship. The images from ImageNet are standard pictures of objects and the main problem is to classify them into many different categories which have nothing to do with ships. However, the neural networks trained on ImageNet somehow capture image features that end up being useful for distinguishing different types of ships. Another interesting point is that transfer learning did not perform well with MSTAR. Understanding these differences could be a valuable research direction. 

This is related to works on distribution shift in simulated-to-real (sim-to-real) scenarios \cite{inkawhich2021bridging, lewis2019sar, arnold2018blending}. The key challenge here is to generate many synthetic training examples which reliably capture the testing distribution (true SAR data). We suspect that our methods could be helpful when trained on synthetic data. In particular, we believe that our methods will perform better when the neural networks are specifically designed to handle SAR data and are trained on that data. These specially designed neural networks should better capture symmetries in the data and the image features will likely be more useful for ATR \cite{9445223}.

\section{CONCLUSION}
In this work, we developed a novel core-set construction method (DAC) and a batch active learning method, LocalMax. Using these methods together with the embedding techniques (CNNVAE and transfer learning) and Laplace learning, these methods beat state-of-the-art SAR classification methods on OpenSARShip and FUSAR-Ship. Additionally, they have dramatically improved efficiency relative to sequential methods, while maintaining the accuracy of sequential methods. These methods also attain higher accuracy than other common batch active learning methods on the datasets tested. 

\section*{ACKNOWLEDGMENTS}
This material is based upon work supported by the 
National Geospatial-Intelligence Agency under Award No.
HM0476-21-1-0003.
Approved for public release, NGA-U-2023-00750.
Any opinions, findings and conclusions or recommendations expressed in this material are those of the authors and do not necessarily reflect the views of the National Geospatial-Intelligence Agency. 
Jeff Calder was also supported by NSF-CCF MoDL+ grant 2212318.

\bibliography{bibliography}
\bibliographystyle{spiebib}

\end{document}